\documentclass[10pt,a4paper,onecolumn]{article}
\usepackage{marginnote}
\usepackage{graphicx}
\usepackage{xcolor}
\usepackage{authblk,etoolbox}
\usepackage{titlesec}
\usepackage{calc}
\usepackage{tikz}
\usepackage{hyperref}
\hypersetup{colorlinks,breaklinks=true,
            urlcolor=[rgb]{0.0, 0.5, 1.0},
            linkcolor=[rgb]{0.0, 0.5, 1.0}}
\usepackage{caption}
\usepackage{tcolorbox}
\usepackage{amssymb,amsmath}
\usepackage{ifxetex,ifluatex}
\usepackage{seqsplit}
\usepackage{xstring}

\usepackage{float}
\let\origfigure\figure
\let\endorigfigure\endfigure
\renewenvironment{figure}[1][2] {
    \expandafter\origfigure\expandafter[H]
} {
    \endorigfigure
}

\usepackage{fixltx2e} 
\usepackage[
  backend=biber,
]{biblatex}
\bibliography{master.bib}

\let\textttOrig=\texttt
\def\texttt#1{\expandafter\textttOrig{\seqsplit{#1}}}
\renewcommand{\seqinsert}{\ifmmode
  \allowbreak
  \else\penalty6000\hspace{0pt plus 0.02em}\fi}

\makeatletter
\let\href@Orig=\href
\def\href@Urllike#1#2{\href@Orig{#1}{\begingroup
    \def\Url@String{#2}\Url@FormatString
    \endgroup}}
\def\href@Notdoi#1#2{\def\tempa{#1}\def\tempb{#2}%
  \ifx\tempa\tempb\relax\href@Urllike{#1}{#2}\else
  \href@Orig{#1}{#2}\fi}
\def\href#1#2{%
  \IfBeginWith{#1}{https://doi.org}%
  {\href@Urllike{#1}{#2}}{\href@Notdoi{#1}{#2}}}
\makeatother

\newlength{\cslhangindent}
\newlength{\csllabelwidth}
\newenvironment{CSLReferences}[2] 
 {
  \setlength{\parindent}{0pt}
  \ifodd #1 \everypar{\setlength{\hangindent}{\cslhangindent}}\ignorespaces\fi
  \ifnum #2 > 0
  \setlength{\parskip}{#2\baselineskip}
  \fi
 }%
 {}
\usepackage{calc}

\usepackage[top=3.5cm, bottom=3cm, right=1.5cm, left=1.7cm,
            headheight=2.2cm, reversemp, includemp, marginparwidth=4.5cm]{geometry}

\titleformat{\section}
  {\normalfont\sffamily\Large\bfseries}
  {}{0pt}{}
\titleformat{\subsection}
  {\normalfont\sffamily\large\bfseries}
  {}{0pt}{}
\titleformat{\subsubsection}
  {\normalfont\sffamily\bfseries}
  {}{0pt}{}
\titleformat*{\paragraph}
  {\sffamily\normalsize}

\usepackage{fancyhdr}
\makeatletter
\let\ps@plain\ps@fancy
\fancyheadoffset[L]{4.5cm}
\fancyfootoffset[L]{4.5cm}

\definecolor{linky}{rgb}{0.0, 0.5, 1.0}

\newtcolorbox{repobox}
   {colback=red, colframe=red!75!black,
     boxrule=0.5pt, arc=2pt, left=6pt, right=6pt, top=3pt, bottom=3pt}

\newcommand{\ExternalLink}{%
   \tikz[x=1.2ex, y=1.2ex, baseline=-0.05ex]{%
       \begin{scope}[x=1ex, y=1ex]
           \clip (-0.1,-0.1)
               --++ (-0, 1.2)
               --++ (0.6, 0)
               --++ (0, -0.6)
               --++ (0.6, 0)
               --++ (0, -1);
           \path[draw,
               line width = 0.5,
               rounded corners=0.5]
               (0,0) rectangle (1,1);
       \end{scope}
       \path[draw, line width = 0.5] (0.5, 0.5)
           -- (1, 1);
       \path[draw, line width = 0.5] (0.6, 1)
           -- (1, 1) -- (1, 0.6);
       }
   }

\patchcmd{\@maketitle}{center}{flushleft}{}{}
\patchcmd{\@maketitle}{center}{flushleft}{}{}
\patchcmd{\@maketitle}{\LARGE}{\LARGE\sffamily}{}{}
\def\maketitle{{%
  
  \AB@maketitle}}
\makeatletter
\renewcommand\AB@affilsepx{ \protect\Affilfont}
\renewcommand\AB@affilnote[1]{{\bfseries #1}\hspace{3pt}}
\renewcommand{\affil}[2][]%
   {\newaffiltrue\let\AB@blk@and\AB@pand
      \if\relax#1\relax\def\AB@note{\AB@thenote}\else\def\AB@note{#1}%
        \setcounter{Maxaffil}{0}\fi
        \begingroup
        \let\href=\href@Orig
        \let\texttt=\textttOrig
        \let\protect\@unexpandable@protect
        \def\thanks{\protect\thanks}\def\footnote{\protect\footnote}%
        \@temptokena=\expandafter{\AB@authors}%
        {\def\\{\protect\\\protect\Affilfont}\xdef\AB@temp{#2}}%
         \xdef\AB@authors{\the\@temptokena\AB@las\AB@au@str
         \protect\\[\affilsep]\protect\Affilfont\AB@temp}%
         \gdef\AB@las{}\gdef\AB@au@str{}%
        {\def\\{, \ignorespaces}\xdef\AB@temp{#2}}%
        \@temptokena=\expandafter{\AB@affillist}%
        \xdef\AB@affillist{\the\@temptokena \AB@affilsep
          \AB@affilnote{\AB@note}\protect\Affilfont\AB@temp}%
      \endgroup
       \let\AB@affilsep\AB@affilsepx
}
\makeatother

\renewcommand\Affilfont{\sffamily\small\mdseries}
\ifnum 0\ifxetex 1\fi\ifluatex 1\fi=0 
  \usepackage[T1]{fontenc}
  \usepackage[utf8]{inputenc}

\else 
  \ifxetex
    \usepackage{mathspec}
    \usepackage{fontspec}

  \else
    \usepackage{fontspec}
  \fi
  \defaultfontfeatures{Ligatures=TeX,Scale=MatchLowercase}

\fi
\IfFileExists{upquote.sty}{\usepackage{upquote}}{}
\IfFileExists{microtype.sty}{%
\usepackage{microtype}
\UseMicrotypeSet[protrusion]{basicmath} 
}{}

\usepackage{hyperref}
\hypersetup{unicode=true,
            pdftitle={sbp-env: Sampling-based Motion Planners'
Testing Environment},
            pdfborder={0 0 0},
            breaklinks=true}
\let\addcontentslineOrig=\addcontentsline
\def\addcontentsline#1#2#3{\bgroup
  \let\texttt=\textttOrig\addcontentslineOrig{#1}{#2}{#3}\egroup}
\let\markbothOrig\markboth
\def\markboth#1#2{\bgroup
  \let\texttt=\textttOrig\markbothOrig{#1}{#2}\egroup}
\let\markrightOrig\markright
\def\markright#1{\bgroup
  \let\texttt=\textttOrig\markrightOrig{#1}\egroup}

\usepackage{graphicx,grffile}
\makeatletter
\def\maxwidth{\ifdim\Gin@nat@width>\linewidth\linewidth\else\Gin@nat@width\fi}
\def\maxheight{\ifdim\Gin@nat@height>\textheight\textheight\else\Gin@nat@height\fi}
\makeatother
\setkeys{Gin}{width=\maxwidth,height=\maxheight,keepaspectratio}
\IfFileExists{parskip.sty}{%
\usepackage{parskip}
}{
\setlength{\parindent}{0pt}
\setlength{\parskip}{6pt plus 2pt minus 1pt}
}
\ifx\paragraph\undefined\else
\let\oldparagraph\paragraph
\renewcommand{\paragraph}[1]{\oldparagraph{#1}\mbox{}}
\fi
\ifx\subparagraph\undefined\else
\let\oldsubparagraph\subparagraph
\renewcommand{\subparagraph}[1]{\oldsubparagraph{#1}\mbox{}}
\fi

\title{sbp-env: Sampling-based Motion Planners'
Testing Environment}

        \author[1]{Tin Lai}

      \affil[1]{School of Computer Science, The University of Sydney,
Australia}
  \date{\vspace{-7ex}}

\begin{document}
\maketitle

\marginpar{

  \begin{flushleft}
  \sffamily\small

  {\bfseries DOI:} \href{https://doi.org/10.21105/joss.03782}{\color{linky}{10.21105/joss.03782}}

  \vspace{2mm}

  {\bfseries Software}
  \begin{itemize}
    \setlength\itemsep{0em}
    \item \href{https://github.com/openjournals/joss-reviews/issues/3782}{\color{linky}{Review}} \ExternalLink
    \item \href{https://github.com/soraxas/sbp-env}{\color{linky}{Repository}} \ExternalLink
    \item \href{https://doi.org/10.5281/zenodo.5572325}{\color{linky}{Archive}} \ExternalLink
  \end{itemize}

  \vspace{2mm}

  \par\noindent\hrulefill\par

  \vspace{2mm}

  {\bfseries Editor:} \href{http://danielskatz.org/}{Daniel S.
Katz} \ExternalLink \\
  \vspace{1mm}
    {\bfseries Reviewers:}
  \begin{itemize}
  \setlength\itemsep{0em}
    \item \href{https://github.com/KanishAnand}{@KanishAnand}
    \item \href{https://github.com/OlgerSiebinga}{@OlgerSiebinga}
    \end{itemize}
    \vspace{2mm}

  {\bfseries Submitted:} 26 September 2021\\
  {\bfseries Published:} 15 October 2021

  \vspace{2mm}
  {\bfseries License}\\
  Authors of papers retain copyright and release the work under a Creative Commons Attribution 4.0 International License (\href{http://creativecommons.org/licenses/by/4.0/}{\color{linky}{CC BY 4.0}}).

  \end{flushleft}
}

\hypertarget{background}{%
\section{Background}\label{background}}

Sampling-based motion planning is one of the fundamental methods by
which robots navigate and integrate with the real world
(\protect\hyperlink{ref-elbanhawi2014_SampRobo}{Elbanhawi \& Simic,
2014}). Motion planning involves planning the trajectories of the
actuated part of the robot, under various constraints, while avoiding
collisions with surrounding obstacles. Sampling-based motion planners
(SBPs) are robust methods that avoid explicitly constructing the often
intractable high-dimensional configuration space (C-Space). Instead,
SBPs randomly sample the C-Space for valid connections and iteratively
build a roadmap of connectivity. Most SBPs are guaranteed to find a
solution if one exists
(\protect\hyperlink{ref-kavraki1996_AnalProb}{Kavraki et al., 1996}),
and such a planner is said to be \emph{probabilistic complete}. A
further development for SBPs is \emph{asymptotic
optimality}(\protect\hyperlink{ref-elbanhawi2014_SampRobo}{Elbanhawi \&
Simic, 2014}): a guarantee that the method will converge, in the limit,
to the optimal solution.

SBPs are applicable to a wide range of applications. Example include
planning with arbitrary cost maps
(\protect\hyperlink{ref-iehlCostmapPlanningHigh2012}{Iehl et al.,
2012}), cooperative multi-agent planning
(\protect\hyperlink{ref-jinmingwuCooperativePathfindingBased2019}{Jiang
\& Wu, 2020}), and planning in dynamic environments
(\protect\hyperlink{ref-yershova2005_DynaRRTs}{Yershova et al., 2005}).
On the one hand, researchers have focused on the algorithmic side of
improving the graph or tree building
(\protect\hyperlink{ref-elbanhawi2014_SampRobo}{Elbanhawi \& Simic,
2014}; \protect\hyperlink{ref-klemmRRTConnectFaster2015}{Klemm et al.,
2015}; \protect\hyperlink{ref-lai2018_BalaGlob}{Lai et al., 2019};
\protect\hyperlink{ref-lai2021rapidlyexploring}{Lai, 2021};
\protect\hyperlink{ref-lai2021lazyExperienceGraph}{Lai \& Ramos, 2021b};
\protect\hyperlink{ref-zhongTripleRrtsRobotPath2012}{Zhong \& Su,
2012}). On the other hand, the advancement of neural networks allows an
abundance of learning approaches to be applied in SBPs
(\protect\hyperlink{ref-bagnell2014_ReinLear}{Bagnell, 2014};
\protect\hyperlink{ref-strubAdaptivelyInformedTrees2020}{Strub \&
Gammell, 2020}) and on improving the sampling distribution
(\protect\hyperlink{ref-alcin2016_ExtrLear}{Alcin et al., 2016};
\protect\hyperlink{ref-lai2020_BayeLoca}{Lai et al., 2020},
\protect\hyperlink{ref-lai2021diffSamp}{2021};
\protect\hyperlink{ref-laiLearningPlanOptimally2020}{Lai \& Ramos,
2020}, \protect\hyperlink{ref-lai2021plannerFlows}{2021a}).

\hypertarget{statement-of-need}{%
\section{Statement of need}\label{statement-of-need}}

The focus of motion planning research has been mainly on (i) the
algorithmic aspect of the planner using different routines to build a
connected graph and (ii) improving the sampling efficiency (with methods
such as heuristic or learned distribution). Traditionally, robotic
research focuses on algorithmic development, which has inspired several
motion planning libraries written in C++, such as Move3D
(\protect\hyperlink{ref-simeon2001move3d}{Simeon et al., 2001}) and OMPL
(\protect\hyperlink{ref-sucan2012open}{Sucan et al., 2012}). In
particular, OMPL has been one of the most well-known motion planning
libraries due to its versatility, and it has been a core part of the
planning algorithm used in the MoveIt framework
(\protect\hyperlink{ref-chitta2012moveit}{Chitta et al., 2012}).
However, swapping the sampler within each planner is very restrictive,
as planners are typically hard-coded to use a specific sampler. In
addition, it is cumbersome to integrate any learning-based approach into
a framework as there is only a limited number of choices of
deep-learning libraries in C++.

Python has been a popular language to use in Machine Learning due to its
rapid scripting nature. For example, PyTorch
(\protect\hyperlink{ref-paszke2019pytorch}{Paszke et al., 2019}) and
Tensorflow (\protect\hyperlink{ref-abadi2016tensorflow}{Abadi et al.,
2016}) are two popular choices for neural network frameworks in Python.
A large number of learning approaches are available as Python packages.
It shall be noted that the aforementioned OMPL has Python bindings
available; however, OMPL uses an outdated Py++ code generator, and every
modification to the source code will require hours to updates bindings
plus recompilation. Some Python repositories are available that are
dedicated to robotics motion planning
(\protect\hyperlink{ref-sakai2018pythonrobotics}{Sakai et al., 2018});
however, most only showcase various planning algorithms, without an
integrated environment and simulators.

\begin{figure}
\centering
\includegraphics{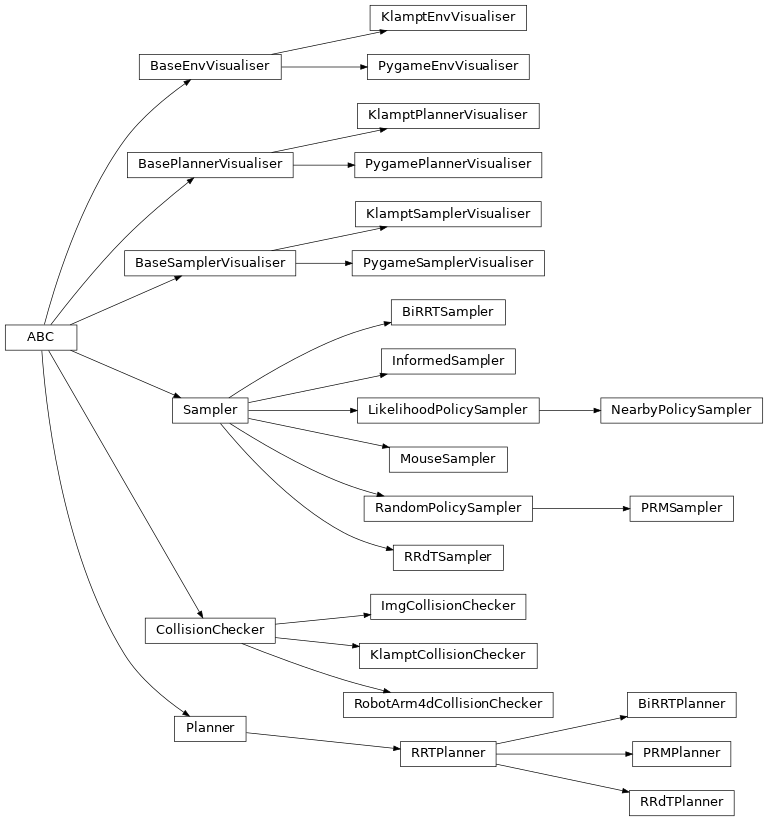}
\caption{Implementation details on the class hierarchy structure of
\texttt{sbp-env}.\label{fig:class-diagram}}
\end{figure}

\hypertarget{overview}{%
\section{Overview}\label{overview}}

We introduce \texttt{sbp-env}, a \emph{sampling-based motion planners'
testing environment}, as a complete feature framework to allow rapid
testing of different sampling-based algorithms for motion planning.
\texttt{sbp-env} focuses on the flexibility of tinkering with different
aspects of the framework, and it divides the main planning components
into two main categories: (i) samplers and (ii) planners. The division
of the two components allows users to decouple them and focus only on
the component that serves as the main focus of the research.
\texttt{sbp-env} has implemented the entire robot planning framework
with multiple degrees-of-freedom, which allows benchmarking motion
planning algorithms with the same planner under different backend
simulators. Separating the two components allows users to quickly swap
out different components in order to test novel ideas.

Building the framework enables researchers to rapidly implement their
novel ideas and validate their hypotheses. In particular, users can
define the environment using something as simple as an \emph{image}, or
as complicated as an \emph{xml file}. All samplers and planners can be
added as a plugin system, and \texttt{sbp-env} will auto-discover newly
implemented planners or samplers that have been added to the dedicated
folders.

Figure \ref{fig:class-diagram} illustrates the hierarical structure of
our package. Our implementation of \texttt{sbp-env} define abstract
interfaces for \textbf{sampler} and \textbf{planners}, from which all
corresponding concrete classes must inherit. In addition, there are
classes that represent the full-body simulations of the environments and
the corresponding visualisation methods. Note that all visualisation can
be turned off on-demand, which is beneficial when users benchmark their
algorithms. The docunmentation of \texttt{sbp-env} is available at
\url{https://cs.tinyiu.com/sbp-env}.

\hypertarget{references}{%
\section*{References}\label{references}}
\addcontentsline{toc}{section}{References}

\hypertarget{refs}{}
\begin{CSLReferences}{1}{0}
\leavevmode\vadjust pre{\hypertarget{ref-abadi2016tensorflow}{}}%
Abadi, M., Barham, P., Chen, J., Chen, Z., Davis, A., Dean, J., Devin,
M., Ghemawat, S., Irving, G., Isard, M.others. (2016). Tensorflow: A
system for large-scale machine learning. \emph{12th {USENIX} Symposium
on Operating Systems Design and Implementation ({OSDI} 16)}, 265--283.

\leavevmode\vadjust pre{\hypertarget{ref-alcin2016_ExtrLear}{}}%
Alcin, O. F., Ucar, F., \& Korkmaz, D. (2016). Extreme learning machine
based robotic arm modeling. \emph{2016 21st International Conference on
Methods and Models in Automation and Robotics, MMAR 2016}, \emph{1},
1160--1163. \url{https://doi.org/gddhmf}

\leavevmode\vadjust pre{\hypertarget{ref-bagnell2014_ReinLear}{}}%
Bagnell, J. A. (2014). Reinforcement {Learning} in {Robotics}: {A
Survey}. \emph{Springer Tracts in Advanced Robotics}, \emph{97}, 9--67.
\url{https://doi.org/gddhk5}

\leavevmode\vadjust pre{\hypertarget{ref-chitta2012moveit}{}}%
Chitta, S., Sucan, I., \& Cousins, S. (2012). Moveit! {[}Ros topics{]}.
\emph{IEEE Robotics \& Automation Magazine}, \emph{19}(1), 18--19.
\url{https://doi.org/10.1109/mra.2011.2181749}

\leavevmode\vadjust pre{\hypertarget{ref-elbanhawi2014_SampRobo}{}}%
Elbanhawi, M., \& Simic, M. (2014). Sampling-based robot motion
planning: {A} review. \emph{IEEE Access}, \emph{2}, 56--77.
\url{https://doi.org/gdkx6g}

\leavevmode\vadjust pre{\hypertarget{ref-iehlCostmapPlanningHigh2012}{}}%
Iehl, R., Cortés, J., \& Siméon, T. (2012). Costmap planning in high
dimensional configuration spaces. \emph{2012 {IEEE}/{ASME International
Conference} on {Advanced Intelligent Mechatronics} ({AIM})}, 166--172.
\url{https://doi.org/gmv7jc}

\leavevmode\vadjust pre{\hypertarget{ref-jinmingwuCooperativePathfindingBased2019}{}}%
Jiang, J., \& Wu, K. (2020). Cooperative pathfinding based on
memory-efficient multi-agent RRT. \emph{IEEE Access}, \emph{8},
168743--168750. \url{https://doi.org/10.1109/access.2020.3023200}

\leavevmode\vadjust pre{\hypertarget{ref-kavraki1996_AnalProb}{}}%
Kavraki, L. E., Kolountzakis, M. N., \& Latombe, J.-C. (1996). Analysis
of probabilistic roadmaps for path planning. \emph{Proceedings of {IEEE
International Conference} on {Robotics} and {Automation}}, \emph{4},
3020--3025. \url{https://doi.org/d66p3t}

\leavevmode\vadjust pre{\hypertarget{ref-klemmRRTConnectFaster2015}{}}%
Klemm, S., Oberländer, J., Hermann, A., Roennau, A., Schamm, T.,
Zollner, J. M., \& Dillmann, R. (2015). {RRT*}-{Connect}: {Faster},
asymptotically optimal motion planning. \emph{2015 {IEEE International
Conference} on {Robotics} and {Biomimetics} ({ROBIO})}, 1670--1677.
\url{https://doi.org/ghpwwm}

\leavevmode\vadjust pre{\hypertarget{ref-lai2021rapidlyexploring}{}}%
Lai, T. (2021). {Rapidly-exploring Random Forest: Adaptively Exploits
Local Structure with Generalised Multi-Trees Motion Planning}.
\emph{arXiv:2103.04487 {[}{c}s.RO{]}}.
\url{https://arxiv.org/abs/2103.04487}

\leavevmode\vadjust pre{\hypertarget{ref-lai2020_BayeLoca}{}}%
Lai, T., Morere, P., Ramos, F., \& Francis, G. (2020). Bayesian {Local
Sampling}-{Based Planning}. \emph{IEEE Robotics and Automation Letters
(RA-L)}, \emph{5}(2), 1954--1961. \url{https://doi.org/gg2n24}

\leavevmode\vadjust pre{\hypertarget{ref-laiLearningPlanOptimally2020}{}}%
Lai, T., \& Ramos, F. (2020). Learning to {Plan Optimally} with
{Flow}-based {Motion Planner}. \emph{arXiv:2010.11323 {[}{c}s.RO{]}}.
\url{https://arxiv.org/abs/2010.11323}

\leavevmode\vadjust pre{\hypertarget{ref-lai2021plannerFlows}{}}%
Lai, T., \& Ramos, F. (2021a). {PlannerFlows: Learning Motion Samplers
with Normalising Flows}. \emph{IEEE/RSJ Proceedings of the International
Conference on Intelligent Robots and Systems (IROS)}.

\leavevmode\vadjust pre{\hypertarget{ref-lai2021lazyExperienceGraph}{}}%
Lai, T., \& Ramos, F. (2021b). {Rapid Replanning in Consecutive
Pick-and-Place Tasks with Lazy Experience Graph}. \emph{arXiv:2109.10209
{[}{c}s.RO{]}}. \url{https://arxiv.org/abs/2109.10209}

\leavevmode\vadjust pre{\hypertarget{ref-lai2018_BalaGlob}{}}%
Lai, T., Ramos, F., \& Francis, G. (2019). Balancing {Global
Exploration} and {Local}-connectivity {Exploitation} with
{Rapidly}-exploring {Random} disjointed-{Trees}. \emph{Proceedings of
{The International Conference} on {Robotics} and {Automation}}.
\url{https://doi.org/ghqdw7}

\leavevmode\vadjust pre{\hypertarget{ref-lai2021diffSamp}{}}%
Lai, T., Zhi, W., Hermans, T., \& Ramos, F. (2021). {Parallelised
Diffeomorphic Sampling-based Motion Planning}. \emph{Conference on Robot
Learning (CoRL)}.

\leavevmode\vadjust pre{\hypertarget{ref-paszke2019pytorch}{}}%
Paszke, A., Gross, S., Massa, F., Lerer, A., Bradbury, J., Chanan, G.,
Killeen, T., Lin, Z., Gimelshein, N., Antiga, L.others. (2019). Pytorch:
An imperative style, high-performance deep learning library.
\emph{Advances in Neural Information Processing Systems}, \emph{32},
8026--8037.

\leavevmode\vadjust pre{\hypertarget{ref-sakai2018pythonrobotics}{}}%
Sakai, A., Ingram, D., Dinius, J., Chawla, K., Raffin, A., \& Paques, A.
(2018). \emph{Python{R}obotics: A {P}ython code collection of robotics
algorithms}. \url{https://arxiv.org/abs/1808.10703}

\leavevmode\vadjust pre{\hypertarget{ref-simeon2001move3d}{}}%
Simeon, T., Laumond, J.-P., \& Lamiraux, F. (2001). Move3D: A generic
platform for path planning. \emph{Proceedings of the 2001 IEEE
International Symposium on Assembly and Task Planning (Isatp2001).
Assembly and Disassembly in the Twenty-First Century.(cat. No.
01th8560)}, 25--30. \url{https://doi.org/10.1109/isatp.2001.928961}

\leavevmode\vadjust pre{\hypertarget{ref-strubAdaptivelyInformedTrees2020}{}}%
Strub, M. P., \& Gammell, J. D. (2020). Adaptively {Informed Trees}
({AIT*}): {Fast Asymptotically Optimal Path Planning} through {Adaptive
Heuristics}. \emph{2020 {IEEE International Conference} on {Robotics}
and {Automation} ({ICRA})}, 3191--3198. \url{https://doi.org/ghvk26}

\leavevmode\vadjust pre{\hypertarget{ref-sucan2012open}{}}%
Sucan, I. A., Moll, M., \& Kavraki, L. E. (2012). The open motion
planning library. \emph{IEEE Robotics \& Automation Magazine},
\emph{19}(4), 72--82.

\leavevmode\vadjust pre{\hypertarget{ref-yershova2005_DynaRRTs}{}}%
Yershova, A., Jaillet, L., Siméon, T., \& LaValle, S. M. (2005).
Dynamic-domain {RRTs}: {Efficient} exploration by controlling the
sampling domain. \emph{Proceedings of {IEEE International Conference} on
{Robotics} and {Automation}}, 3856--3861.
\url{https://doi.org/10.1109/robot.2005.1570709}

\leavevmode\vadjust pre{\hypertarget{ref-zhongTripleRrtsRobotPath2012}{}}%
Zhong, J., \& Su, J. (2012). Triple-{Rrts} for robot path planning based
on narrow passage identification. \emph{2012 {International Conference}
on {Computer Science} and {Information Processing} ({CSIP})}, 188--192.
\url{https://doi.org/ghjdnk}

\end{CSLReferences}

\end{document}